\documentclass[10pt,letterpaper]{article}

\usepackage{cogsci}

\twopagesummarysubmission 

\usepackage[hyphens]{url}
\usepackage{pslatex}
\usepackage{hyperref}
\usepackage{apacite}
\usepackage{float} 

\usepackage[none]{hyphenat} 

\usepackage{amsmath}
\usepackage{amssymb}
\usepackage{graphicx}
\usepackage{nicefrac}

\title{Biological Plausibility and Representational Alignment of Feedback Alignment in Convolutional Networks}
 
\author{%
  {\large \bf Jake Lance\textsuperscript{*} (jake.lance@mail.utoronto.ca)} \\
  University of Toronto
  \AND
  {\large \bf Larry Kieu\textsuperscript{*} (kieu@cs.toronto.edu)} \\
  University of Toronto
}

\date{} 

\begin{document}
\maketitle

\begingroup
\renewcommand\thefootnote{*}
\footnotetext{Equal contribution.}
\endgroup

\begin{abstract}

The feedback alignment (FA) algorithm offers a biologically plausible alternative to backpropagation (BP) for training neural networks yet notably fails to scale to convolutional architectures. Modifications have been proposed to address this limitation, but at questionable cost to biological plausibility. In this paper, we evaluate five learning algorithms including modified FA and standard BP, applied to the same convolutional architecture with the CIFAR-10 dataset. We provide a tripartite comparative analysis focusing on biological plausibility, interpretability, and computational complexity. Our results indicate that modified FA algorithms converge on internal representations that are structurally similar to those produced by backpropagation. In particular, it appears the functional success of modified FA algorithms may be rooted in their ability to mimic the representational geometry of backpropagation, converging on similar representations despite relying on fundamentally different weight update mechanisms. All codes to reproduce our experiments can be found \href{https://github.com/dqieu/Interp_FA}{here}.

\end{abstract}

\section{Introduction}

Backpropagation (BP) remains the dominant algorithm for training deep neural networks, achieving state-of-the-art performance across vision, language, and reinforcement learning tasks. At its core, BP computes error gradients by propagating a loss signal backward through a network using the transpose of the forward weight matrices. While mathematically elegant, this requirement, known as the weight transport problem, \shortciteA{grossberg_competitive_1987}, has received scrutiny from computational neuroscientists for being biologically implausible. In particular, this requirement demands that each feedback synapse carry information precisely mirroring its corresponding feedforward synapse. No known biological mechanism supports this kind of symmetric synaptic communication, making BP difficult to reconcile with what we know about learning in the brain.

Feedback alignment (FA) was introduced by \shortciteA{lillicrap_random_2016} as a biologically motivated alternative. Rather than propagating errors through the transposed forward weights, FA uses a fixed, random feedback matrix \emph{B} to convey the error signal backward. Remarkably, networks trained this way still learn. The forward weights gradually align with the random feedback weights such that the FA gradient becomes correlated with the true BP gradient, a phenomenon termed \textit{gradient alignment}. This result attracted substantial attention from both the machine learning and neuroscience communities, as it suggested that exact weight symmetry might not be necessary for credit assignment in deep networks. 
\footnote{The significance of this contribution continues to be recognized, including the most recent 2025 Hinton-Sejnowski Prize at NeurIPS (\href{https://blog.neurips.cc/2025/11/26/announcing-the-2025-sejnowski-hinton-prize/}{link}), for its ``significant impact'' and ``helping to establish a new sub-field of `biologically plausible' learning rules in the NeurIPS community and beyond''.}

Despite the impressive theoretical contributions, FA's promise comes with a critical limitation. While competitive with BP in shallow fully-connected networks, traditional FA fails to scale to convolutional architectures and more complex datasets \shortciteA{bartunov_assessing_2018, liao2016importantweightsymmetrybackpropagation}. The underlying problem is twofold: as network depth increases, the misalignment between the fixed feedback matrix and the changing forward weights causes the gradient signal to vanish or explode exponentially across layers, and the FA gradient angle with respect to the true BP gradient fails to converge in lower layers. These failure modes are manageable in shallow networks but become critical at depth, limiting FA's applicability to large architectures characteristic of modern deep learning.
Despite the impressive theoretical contributions, FA's promise comes with a critical limitation. While competitive with BP in shallow fully-connected networks, traditional FA fails to scale to convolutional architectures and more complex datasets \shortciteA{bartunov_assessing_2018, liao2016importantweightsymmetrybackpropagation}. The underlying problem is twofold: as network depth increases, the misalignment between the fixed feedback matrix and the changing forward weights causes the gradient signal to vanish or explode exponentially across layers, and the FA gradient angle with respect to the true BP gradient fails to converge in lower layers. These failure modes are manageable in shallow networks but become critical at depth, limiting FA's applicability to large architectures characteristic of modern deep learning.
Despite the impressive theoretical contributions, FA's promise comes with a critical limitation. While competitive with BP in shallow fully-connected networks, traditional FA fails to scale to convolutional architectures and more complex datasets \shortciteA{bartunov_assessing_2018, liao2016importantweightsymmetrybackpropagation}. The underlying problem is twofold: as network depth increases, the misalignment between the fixed feedback matrix and the changing forward weights causes the gradient signal to vanish or explode exponentially across layers, and the FA gradient angle with respect to the true BP gradient fails to converge in lower layers. These failure modes are manageable in shallow networks but become critical at depth, limiting FA's applicability to the large architectures characteristic of modern deep learning.

To address these limitations, several modifications to the FA algorithm have been proposed. The most influential, introduced by \shortciteA{liao2016importantweightsymmetrybackpropagation} and extended by \shortciteA{moskovitz_feedback_2018}, is uniform sign-concordant feedback (uSF); rather than keeping \textit{B} entirely fixed and random, the feedback weights are updated to track the sign of the corresponding forward weights. Two variants of this approach have been studied, including the initialization method (uSF Init), in which \textit{B} is scaled by the magnitude of the initial feedback weights; and the strict normalization method (uSF SN), in which \textit{B} additionally tracks the relative magnitude of the forward weights. Both methods substantially improve gradient alignment and recover performance competitiveness with BP on CIFAR-10 and ImageNet. However, as we examine in this paper, these modifications do so by progressively reintroducing dependencies between the feedback and forward weights, quietly walking back the biological motivation that made FA compelling in the first place.

In this work, we ask a question the performance-focused FA literature has not addressed: do modifications that improve gradient alignment with BP also cause networks to converge on more BP-like internal representations? We additionally ask if achieving this convergence requires sacrificing biological plausibility? To investigate this, we train the same CNN architecture on CIFAR-10 using BP, standard FA, uSF Init, and uSF SN, and analyze the resulting networks using centered kernel alignment (CKA) to measure representational similarity and layer-wise gradient alignment to characterize the learning process. Together, these measures allow us to assess whether the plausibility-performance tradeoff in modified FA algorithms has a corresponding representational signature and what that signature tells us about the limits of biologically plausible learning in deep networks.
\begin{figure}[H]
\centering
\includegraphics[width=\columnwidth]{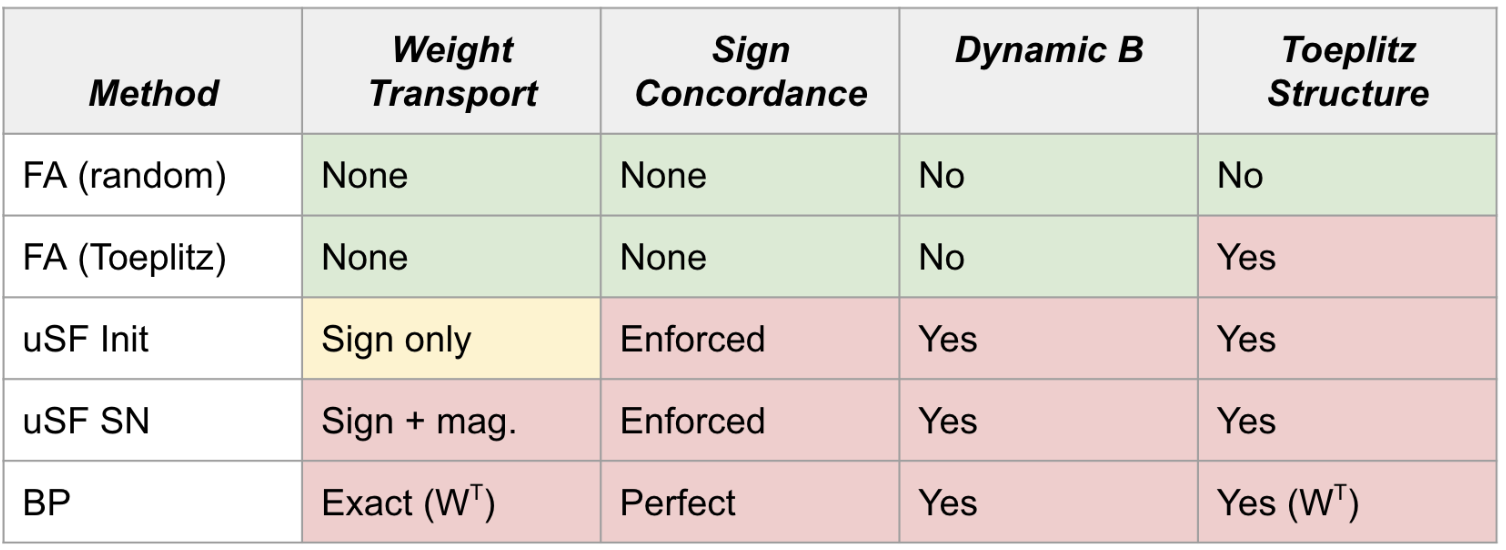}
\caption{Biological Plausibility taxonomy of learning algorithms. Green indicates higher biological plausibility, yellow for decreased plausibility, and red for implausibility.}
\label{fig:biodiff}
\end{figure}

\section{Materials and Methods}
\label{sec:methods}

\subsection{Learning Algorithms}
\label{sec:learning_algorithms}

We compare five learning rules that differ only in how the error signal is propagated backward through the network. In all cases, the \emph{weight update} at each layer is computed from the local error signal and the pre-synaptic activations in the standard way; the methods differ solely in how that error signal is obtained.

\paragraph{Backpropagation (BP).}
Standard backpropagation computes the error signal $\boldsymbol{\delta}^{l} = \partial \mathcal{L} / \partial \mathbf{u}^{l}$ at layer $l$ by transposing the forward weight matrix:
\begin{equation}
    \boldsymbol{\delta}^{l} = \left({W^{l+1}}^\top \boldsymbol{\delta}^{l+1}\right) \odot f'\!\left(\mathbf{u}^{l}\right),
    \label{eq:bp}
\end{equation}
where $\mathcal{L}$ is the loss, $\mathbf{u}^{l}$ is the pre-activation at layer $l$, $f'$ is the derivative of the activation function, and $\odot$ denotes the element-wise product. For convolutional layers, the transpose operation corresponds to a convolution with the 180$^{\circ}$-rotated forward kernel $\bar{W}^{l+1}$:
\begin{equation}
    \boldsymbol{\delta}^{l}_{i} = \left(\bar{W}^{l+1}_{i} * \boldsymbol{\delta}^{l+1}_{i}\right) \odot f'\!\left(\mathbf{u}^{l}_{i}\right),
    \label{eq:bp_conv}
\end{equation}
where $*$ denotes convolution and $i$ indexes channels. This requires the backward path to have exact knowledge of the forward weights---the ``weight transport problem''~\shortcite{grossberg_competitive_1987, lillicrap_random_2016}.

\paragraph{Feedback Alignment (FA).}
Feedback Alignment \shortcite{lillicrap_random_2016} replaces the transposed forward weights with a random feedback matrix $B$ that is fixed at the beginning of training:
\begin{equation}
    \boldsymbol{\delta}^{l} = \left(B^{l+1}\, \boldsymbol{\delta}^{l+1}\right) \odot f'\!\left(\mathbf{u}^{l}\right).
    \label{eq:fa}
\end{equation}
The weight update rule remains identical to BP.

\paragraph{FA in convolutional layers: Dense vs.\ Toeplitz feedback.}
The original FA formulation \shortcite{lillicrap_random_2016} was developed for fully-connected networks. Extending FA to convolutional architectures, we identify two natural forms:

\begin{description}
    \item[FA (Random)] \emph{Dense matrix multiplication.} The feedback is applied as a dense matrix--vector product between a fixed random matrix $B \in \mathbb{R}^{d_{\text{out}} \times d_{\text{in}}}$ and the incoming error signal. This treats the convolutional layer's backward pass as if it were fully-connected, discarding the spatial structure of the convolution entirely.

    \item[FA (Toeplitz)] \emph{Convolutional feedback.} The feedback is applied as a transposed convolution with a fixed random kernel $B_0 \in \mathbb{R}^{C_{\text{out}} \times C_{\text{in}} \times k_H \times k_W}$:
    \begin{equation}
        \tilde{\boldsymbol{\delta}}^{l} = B_0 *^{\!\top} \boldsymbol{\delta}^{l+1},
        \label{eq:fa_toeplitz}
    \end{equation}
    where $*^{\!\top}$ denotes the transposed convolution (with the same padding and stride as the forward layer).
\end{description}

\paragraph{Uniform sign-concordant feedback (uSF) variants.}
\shortciteA{refinetti_align_2021} identified angular alignment between the FA error signal and the true BP gradient as a common feature of successful FA learning, showing that networks first align their weights with the feedback before memorizing the data. Building on this, \shortciteA{moskovitz_feedback_2018} proposed enforcing \emph{sign concordance} between the feedback and forward weights---a lightweight form of alignment---to close the performance gap in deep convolutional networks:

\begin{description}
    \item[uSF Init:] The feedback matrix at training step $t$ is
    \begin{equation}
        B^{l}_t = \left|B^{l}_0\right| \odot \operatorname{sign}\!\left(W^{l}_t\right),
        \label{eq:usf_init}
    \end{equation}
    where $B^{l}_0$ is the initial random feedback matrix. This retains the random magnitudes from initialization but copies the sign pattern of the current forward weights at each step.

    \item[uSF SN (Strict Normalization):] The feedback is
    \begin{equation}
        B^{l}_t = \|W^{l}_t\|_2 \frac{\operatorname{sign}\!\left(W^{l}_t\right)}{\|\operatorname{sign}\!\left(W^{l}_t\right)\|_2},
        \label{eq:usf_sn}
    \end{equation}
    which normalizes the feedback to match the spectral norm of the forward weights while using only sign information. Notably, $B^l_t$ under uSF~SN is entirely determined by $W^l_t$---the initial random matrix $B^l_0$ plays no role.
\end{description}

Both uSF variants use the Toeplitz (convolutional) feedback structure for convolutional layers.

\paragraph{Implementation details.}
 At the start of training, the feedback matrices $B_0$ are drawn from $\mathcal{N}(0, \sigma^2 I)$ where $\sigma^2$ is a scaling factor. We adopt Architecture~1 from \shortciteA{moskovitz_feedback_2018} and train on CIFAR-10 \shortcite{krizhevsky_learning_2009} with the Adam optimizer \shortcite{kingma_adam_2015} until convergence. For comparability, we set a common global seed. Full details are provided in the Supplementary Information.

\subsection{Gradient Alignment and Sign Concordance}
\label{sec:alignment}

\paragraph{Gradient alignment.}
A feature of successful learning with FA is that the forward weights gradually align with the fixed feedback matrix during learning \shortcite{lillicrap_random_2016}. We track this alignment by computing the angle between the true BP gradient $\partial \mathcal{L} / \partial \mathbf{u}^{l}$ and the FA feedback signal $\tilde{\boldsymbol{\delta}}^{l}$.

\paragraph{Sign concordance.}
We measure the fraction of weight entries where $\operatorname{sign}(W^l_t) = \operatorname{sign}(B^l_0)$. Since the uSF methods enforce sign agreement by construction, we track this metric only for the unmodified FA variants (FA Random and FA Toeplitz) to investigate whether sign concordance emerges naturally alongside angular alignment.

\subsection{Centered Kernel Alignment (CKA)}
\label{sec:cka}

To compare the internal representations learned by different methods, we use linear Centered Kernel Alignment (CKA) \shortcite{kornblith_similarity_2019}, which measures similarity between two sets of neural activations in a way that is invariant to orthogonal transformations and isotropic scaling. Our implementation follows \shortciteA{nguyen_do_2021}. For each FA variant, we compute the layer-to-layer CKA matrix between the BP model and the FA model. From the untrained data, we also evaluate CKA on the subset of samples where both BP and the FA method classify correctly, and samples where BP succeeds but the FA method fails---where representations can most diverge.

\subsection{Feature Visualization}
\label{sec:feature_vis}

To understand \emph{what} each method's convolutional filters learn to detect, we perform a targeted feature visualization analysis. Rather than examining all channels, we design a pipeline that identifies the channels most relevant to a particular classification decision, then characterizes those channels by the real images that maximally activate them.

\paragraph{Focusing on dogs.}
Preliminary failure analysis revealed that the ``dog'' class is the category where FA (Random) and FA (Toeplitz) exhibit their largest accuracy deficit relative to BP. We therefore focus our feature visualization on the dog class to understand what these methods fail to learn about dog-discriminative features.

\paragraph{Identifying influential channels via GradCAM-style importance.}
For each method $m$ and each convolutional layer $\ell \in \{\text{conv1}, \text{conv2}\}$, we compute a per-channel importance score adapted from Grad-CAM \shortcite{selvaraju_grad-cam_2020}. For each dog image $\mathbf{x}_i$ in the test set, we perform a forward pass, recording the activation $A_c^{(\ell)} \in \mathbb{R}^{H \times W}$ at channel $c$ of layer $\ell$. We then backpropagate from the model's own predicted class $\hat{y}_i = \arg\max_k f_k(\mathbf{x}_i)$ (not necessarily the ground-truth label) to obtain the gradient $\frac{\partial f_{\hat{y}_i}}{\partial A_c^{(\ell)}}$. The importance of channel $c$ for image $i$ is:
\begin{equation}
    \alpha_{c,i}^{(\ell)} = \left|\,\overline{g}_{c,i}^{(\ell)} \cdot \overline{A}_{c,i}^{(\ell)}\,\right|,
    \label{eq:channel_importance}
\end{equation}
where $\overline{g}_{c,i}^{(\ell)}$ and $\overline{A}_{c,i}^{(\ell)}$ are the spatial means of the gradient and activation maps, respectively. We average this importance over all 1{,}000 dog test images and select the top 3 channels per layer for detailed visualization.

Note that all models are loaded with their learned weights into a standard BP architecture, so that the gradient flow used for importance computation is identical across methods---only the learned weights differ.

\paragraph{Top-activating image exemplars.}
For each of the top-3 channels, we scan the full test set (10{,}000 images) and find the 9 images that produce the highest mean spatial activation for that channel. By examining which images maximally excite a channel, we obtain an interpretable description of what visual features the channel has learned to detect.

\subsection{Computational Efficiency}
\label{sec:efficiency}

The FA variants introduce different computational overheads relative to BP. FA~(Toeplitz) is roughly equivalent to BP in cost, since the backward pass performs a transposed convolution with a fixed kernel of the same shape as the forward kernel. FA~(Random), by contrast, is the least efficient: its dense feedback matrix for convolutional layers has $d_{\text{out}} \times d_{\text{in}}$ parameters (compared to $C_{\text{out}} \times C_{\text{in}} \times k_H \times k_W$ for the Toeplitz kernel), making its backward pass substantially more expensive in both computation and memory. The uSF variants incur an additional $O(|W|)$ cost per backward pass to recompute the sign-flipped feedback from the current weights, which is modest relative to the forward and backward pass costs. However, uSF~Init additionally requires storing the initial random matrix $B_0$ throughout training, adding $O(|W|)$ extra memory per layer.

\section{Results}
\label{sec:results}

\subsection{Training Accuracy}

\begin{figure}[H]
\centering
\includegraphics[width=.8\columnwidth]{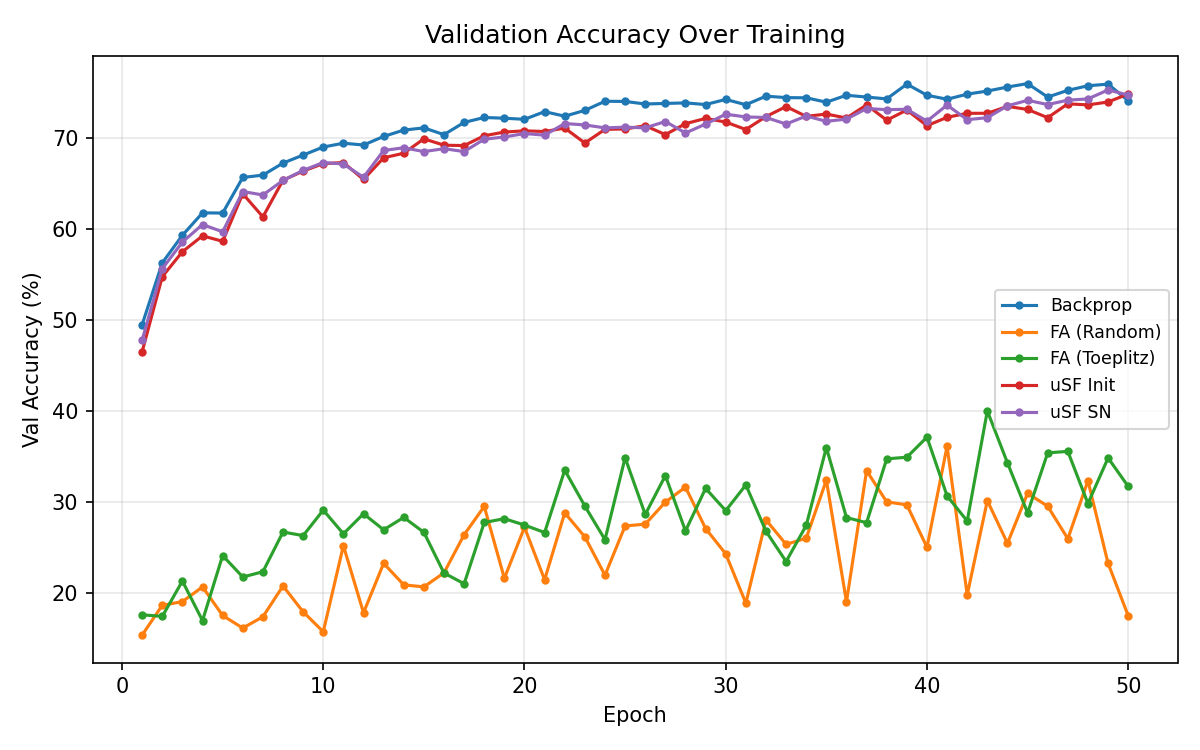}
\caption{Validation accuracy over 50 epochs of training.}
\label{fig:val_acc}
\end{figure}

Table~\ref{tab:accuracy} reports test accuracy for each method, reproducing prior literature \shortciteA{moskovitz_feedback_2018}. Figure~\ref{fig:val_acc} shows the validation accuracy trajectories: BP and the uSF variants converge to similar levels within 10--15 epochs, while the pure FA methods plateau early at much lower accuracy.

\begin{table}[H]
\centering
\begin{tabular}{lcc}
    \hline
    \textbf{Method} & \textbf{Best Val (\%)} & \textbf{Test (\%)} \\
    \hline
    BP              & 75.92 & 75.80 \\
    FA (Random)     & 36.14 & 35.77 \\
    FA (Toeplitz)   & 40.02 & 40.07 \\
    uSF Init        & 74.78 & 74.30 \\
    uSF SN          & 75.22 & 75.67 \\
    \hline
\end{tabular}
\caption{Validation and test accuracy on CIFAR-10}
\label{tab:accuracy}
\end{table}

\subsection{Gradient Alignment}

\begin{figure}[H]
\centering
\includegraphics[width=.9\columnwidth]{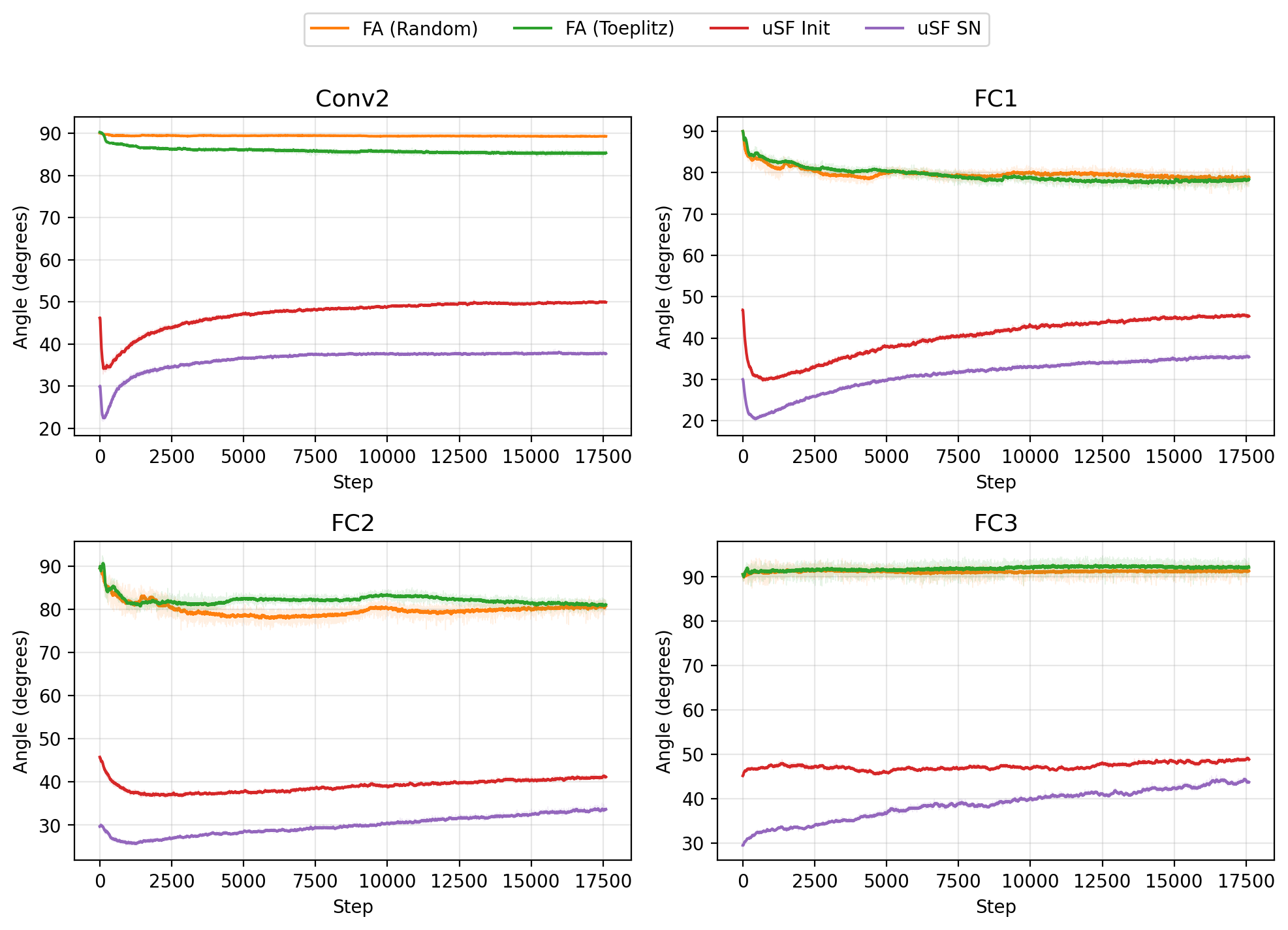}
\caption{Angle between the BP gradient and the feedback signal during training. Lower angles indicate better alignment.}
\label{fig:alignment}
\end{figure}

Figure~\ref{fig:alignment} tracks the angle between the true BP gradient and each method's feedback signal during training. Over training, the uSF variants rapidly decrease to substantially lower angles, indicating strong alignment with the BP gradient. Consistent with the dynamics described by \shortciteA{refinetti_align_2021}, the alignment then plateaus or rises again as the model prioritizes data fitness. The pure FA methods also improve, but their angles remain considerably higher, particularly in the convolutional layers. This persistent misalignment in the convolutional layers may partly explain the large accuracy gap between the pure FA variants and the uSF methods.

\subsection{Sign Concordance}

\begin{figure}[H]
\centering
\includegraphics[width=\columnwidth]{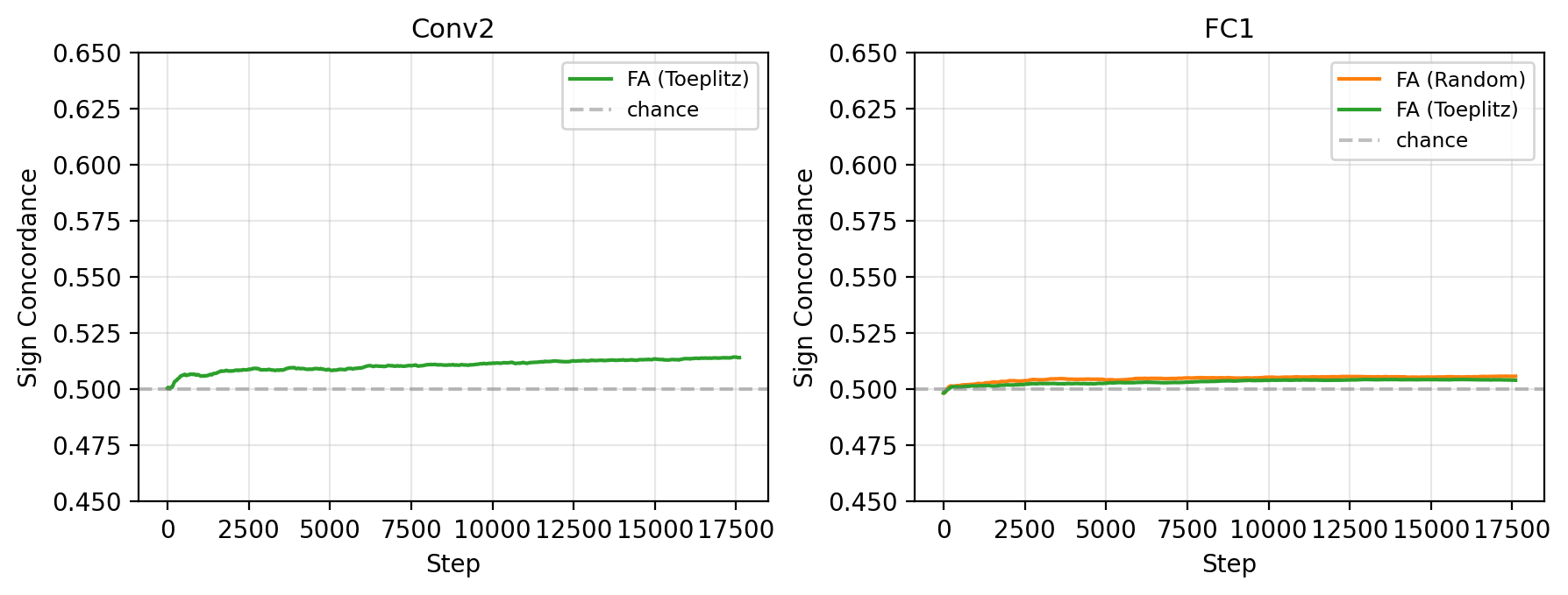}
\caption{Sign concordance (fraction of elements where $\operatorname{sign}(W) = \operatorname{sign}(B)$) during training for FA~(Toeplitz). The dashed line indicates the 0.5 chance level. Note: FA~(Random)'s dense feedback matrix has a different shape from the convolutional weights, precluding element-wise sign comparison. Sign concordance results for other layers are included in Supplementary Information.}
\label{fig:sign}
\end{figure}

Figure~\ref{fig:sign} shows sign concordance between the forward weights and the fixed feedback matrices for the unmodified FA variants. Across all layers and both methods, sign concordance remains near the chance level of 0.5 throughout training. Despite improvements in angular alignment, the forward weights do not develop meaningful sign agreement with the random feedback matrices. This is consistent with the poor performance of the unmodified FA variants and supports the rationale for enforcing sign agreement in the uSF variants.

\subsection{Centered Kernel Alignment}

\begin{figure}[H]
\centering
\includegraphics[width=0.49\columnwidth]{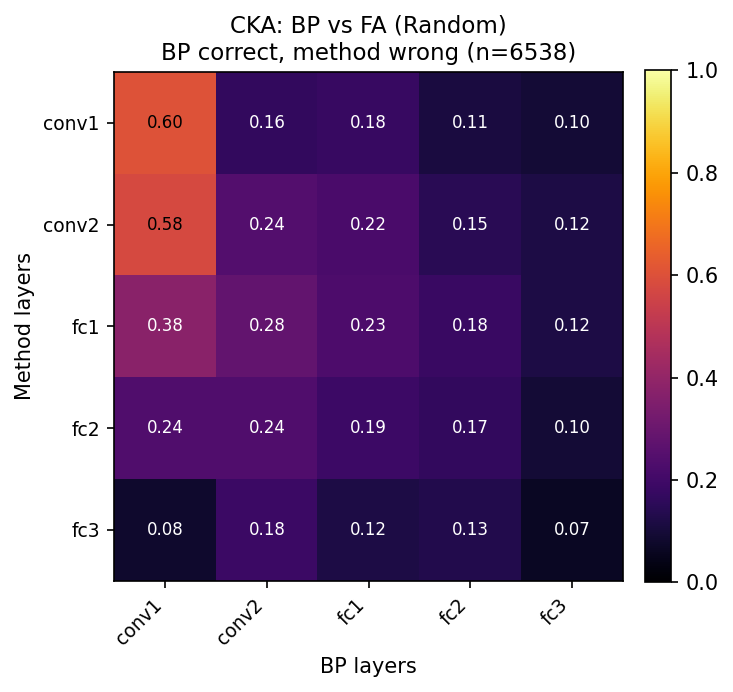}
\includegraphics[width=0.49\columnwidth]{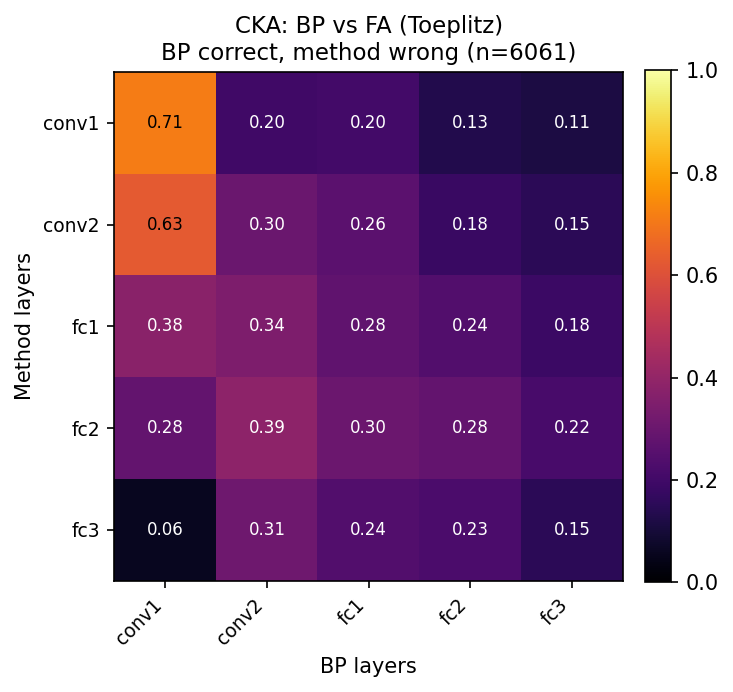}\\[2pt]
\includegraphics[width=0.49\columnwidth]{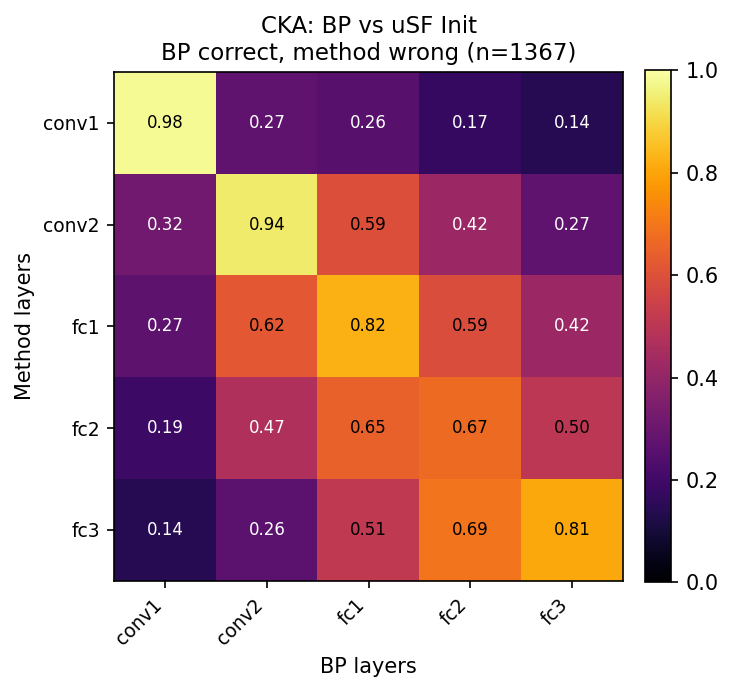}
\includegraphics[width=0.49\columnwidth]{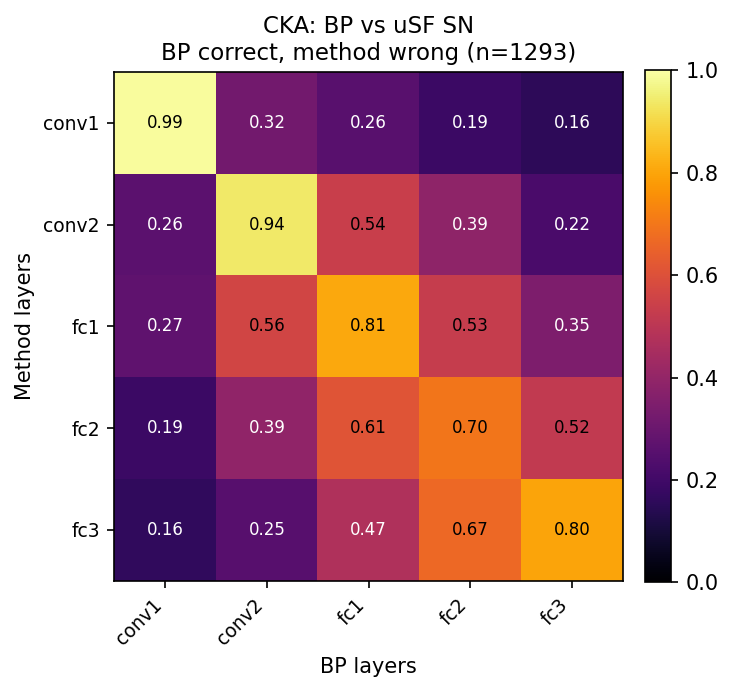}
\caption{CKA similarity between BP and each FA variant, evaluated on images where BP is correct but the FA method fails.}
\label{fig:cka}
\end{figure}

Figure~\ref{fig:cka} shows CKA heatmaps comparing each FA variant's layer-wise representations to BP, evaluated on the subset of images where BP classifies correctly but the FA method fails.

For FA~(Random) and FA~(Toeplitz), CKA values are uniformly low, even along the diagonal. This confirms that these methods learn fundamentally different representations from BP at every layer, particularly in the deeper layers where representations collapse toward near-zero similarity. An off-diagonal pattern reinforces this: FA~(Random)'s second layer (conv2) representations are most similar to BP's first layer (conv1) rather than BP's second layer, and similarly for FA~(Toeplitz). This suggests that the pure FA variants fail to learn meaningful additional features beyond what the first layer already captures---stalling at the first level of representation.

The uSF variants present a strikingly different picture. Both uSF~Init and uSF~SN maintain high diagonal CKA values, indicating that even on misclassified images, these methods learn nearly identical early representations to BP. Similarity in deeper layers remains substantial, suggesting that classification failures in uSF methods stem from subtle differences in the decision layers rather than fundamentally different feature extraction. CKA results on the full validation dataset and on the subset where both methods classify correctly are provided in the Supplementary Information.

\subsection{Feature Visualization}

\begin{figure}[H]
\centering
\includegraphics[width=.8\columnwidth]{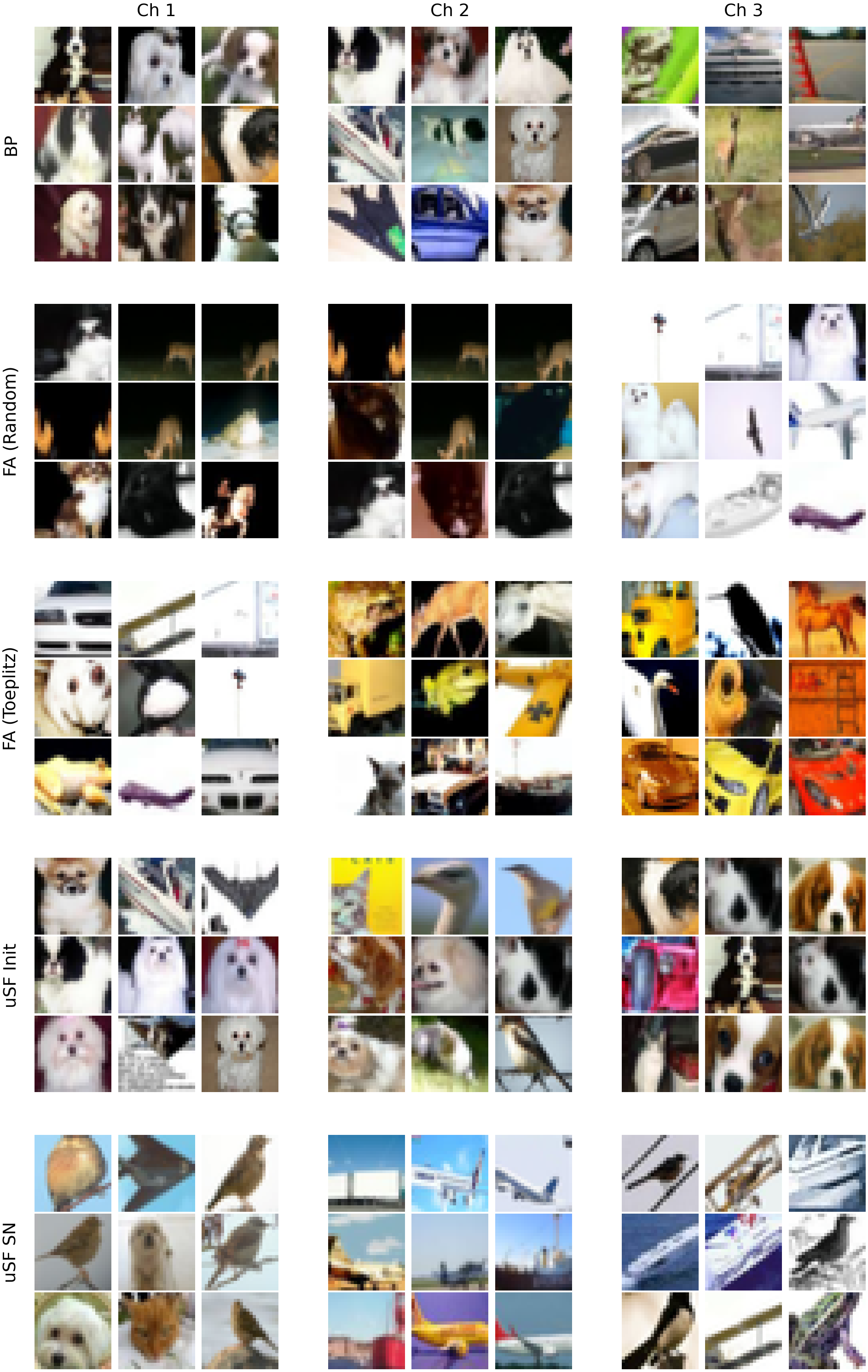}
\caption{Top-activating test images for the three most dog-important Conv2 channels per method. Each row block is one method (labelled at left); each column block is one channel. Within each block, the 9 highest-activating images are arranged in a 3$\times$3 grid.}
\label{fig:exemplars}
\end{figure}

Figure~\ref{fig:exemplars} shows the top-activating test images for the three most important channels of the last convolutional layer (identified via GradCAM-style importance on dog images) for each method. Each row corresponds to one channel, and each column shows one of the nine images that most strongly activate it.

BP's top channels are activated by images of small white dogs (row~1) and dogs with distinctive fur patterns (row~2), demonstrating that BP learns dog-discriminative features at this layer. uSF~Init shows a similar pattern: its top channels respond to small dogs and puppies, consistent with the high CKA similarity to BP observed above.

FA~(Random)'s top channels are activated by dark, low-contrast images---scenes with dark backgrounds rather than any identifiable object features. This suggests that FA~(Random) relies on low-level luminance statistics rather than learning semantic, class-discriminative features in Conv2. FA~(Toeplitz) shows slightly more varied activations (responding to bright objects, animals, and vehicles), but without clear dog specificity.

uSF~SN's top channels respond to birds and elongated objects (ships, planes), rather than dogs. Despite achieving near-BP accuracy overall, uSF~SN appears to route dog classification through different feature pathways than BP.

\section{Discussion}

Together, our CKA and biological plausibility analysis reveals an interesting pattern: the degree to which a method converges on BP-like internal representations appears to track closely with how much biological plausibility it sacrifices. The uSF methods show near-perfect representational agreement with BP in the early convolutional layers, with uSF Init and uSF SN both achieving CKA scores of 0.99 at conv1, and 0.94 and 0.93 at conv2 respectively, suggesting that the first two layers have converged on essentially identical feature representations to those learned by backpropagation. This agreement degrades across the fully-connected layers, where scores fall to the 0.75–0.87 range, indicating that while early feature extraction is nearly identical to BP, the deeper classification layers retain more method-specific structure. The FA variants tell an opposing story, albeit one rooted in their known poor performance on convolutional architectures. FA (Toeplitz) achieves a diagonal CKA of 0.72 at conv1, dropping to 0.06 at fc3, with the network's representations diverging substantially from BP as depth increases. FA (Random) is weaker still, with conv1 at 0.60 and fc3 at 0.09. Neither FA variant produces a well-conditioned diagonal structure in the off-diagonal CKA cells, which is expected, given the known limitation of standard FA to learn on convolutional architectures. 

These results map directly onto the biological plausibility taxonomy summarized in Figure~\ref{fig:biodiff}. Reading down the first column---FA (Random), FA (Toeplitz), uSF Init, uSF SN---each algorithm in this order introduces an additional constraint, while giving rise to more BP-like representations. For example, standard FA requires no knowledge of W and learns the least BP-like solution. The uSF methods enforce sign concordance and introduce dynamic updating of B, and achieve near-perfect early-layer convergence with BP.

Several limitations constrain the scope of our conclusions. First, all experiments use a single five-layer CNN architecture trained on CIFAR-10; it is not known whether the representational convergence patterns observed here hold for deeper architectures or more complex datasets. Second, CKA is sensitive to the choice of comparison set and layer granularity, with the layer-by-layer picture we report reflecting one reasonable set of design choices; finer-grained analysis might reveal within-layer variation not captured here. Finally, the results reported reflect a limited number of training runs; additional replication would strengthen confidence in the stability of the observed CKA values, particularly for the FA variants where variance may be higher.

\section{Conclusion}

This paper set out to ask a question the performance-focused feedback alignment literature has not appeared to have addressed: do the modifications that make FA scale to convolutional architectures cause networks to converge on more BP-like internal representations, and does achieving that convergence sacrifice biological plausibility? Our CKA analysis across five learning algorithms trained on CIFAR-10 provides a clear answer to both. The uSF methods which enforce sign concordance between feedback and forward weights and require dynamic updating of \textit{B} achieve near-perfect representational agreement with BP in early layers, while standard FA variants, which require no knowledge of W, learn structurally different internal solutions. The degree of representational convergence tracks the degree of biological plausibility cost, layer by layer. The implication is pointed: the modifications that make feedback alignment work in deep convolutional neural networks do so by making it progressively less like feedback alignment and more like backpropagation, in both the learning signal and the representations it produces. Each modification seems to improve performance by reintroducing a dependency the algorithm was originally notable for avoiding, such that what remains of FA's biological motivation after modification is largely surface-level. Overall, the functional success of these algorithms appears rooted in their approximation of BP's solution, not in any genuinely novel credit assignment strategy. Whether biologically plausible learning algorithms can find solutions dissimilar to backpropagation at competitive performance remains an open question and one we suggest is worth exploring.

\bibliographystyle{apacite}

\setlength{\bibleftmargin}{.125in}
\setlength{\bibindent}{-\bibleftmargin}

\bibliography{ProposalCOG403Proj}
\section{Appendix}
\label{sec:supplementary}

\subsection{Architecture}
\label{sec:supp_architecture}

We adopt Architecture~1 from \shortciteA{moskovitz_feedback_2018}, a convolutional network designed for $24 \times 24 \times 3$ inputs:

\begin{center}
\begin{tabular}{ll}
    \hline
    \textbf{Layer} & \textbf{Configuration} \\
    \hline
    Conv1 + ReLU & $3 \to 64$, $5 \times 5$ \\
    MaxPool & $2 \times 2$ \\
    Conv2 + ReLU & $64 \to 64$, $5 \times 5$ \\
    MaxPool & $2 \times 2$ \\
    FC1 + ReLU & $576 \to 384$ \\
    FC2 + ReLU & $384 \to 192$ \\
    FC3 & $192 \to 10$ \\
    \hline
\end{tabular}
\end{center}

The model is trained with cross-entropy loss.

\subsection{Dataset}
\label{sec:supp_dataset}

We use CIFAR-10 \shortcite{krizhevsky_learning_2009}, comprising 60{,}000 $32 \times 32$ color images in 10 classes. To match the network's $24 \times 24$ input size, training images are randomly cropped to $24 \times 24$ and randomly flipped horizontally. Validation and test images are center-cropped to $24 \times 24$. All images are normalized per-channel (mean $= (0.4914, 0.4822, 0.4465)$, std $= (0.2470, 0.2435, 0.2616)$).

The 50{,}000 training images are split 90/10 into 45{,}000 training and 5{,}000 validation samples (fixed seed $= 42$). The held-out 10{,}000 test images are used only for final evaluation.

\subsection{Hyperparameter Search}
\label{sec:supp_hpsearch}

For each method, we perform a grid search, training for 3 epochs and selecting the configuration with the highest validation accuracy:
\begin{itemize}
    \item \textbf{BP}: learning rate $\in \{5 \times 10^{-4},\; 10^{-3},\; 3 \times 10^{-3}\}$, weight decay $\in \{0,\; 10^{-4}\}$.
    \item \textbf{FA, FA Toeplitz, uSF Init}: learning rate $\in \{10^{-3},\; 3 \times 10^{-3},\; 10^{-2}\}$, feedback scale $\sigma \in \{0.01,\; 0.05,\; 0.1\}$.
    \item \textbf{uSF SN}: learning rate $\in \{10^{-3},\; 3 \times 10^{-3},\; 10^{-2},\; 3 \times 10^{-2}\}$ (feedback scale is irrelevant since $B^l_t$ is fully determined by $W^l_t$).
\end{itemize}

All methods are trained with Adam ($\beta_1 = 0.9$, $\beta_2 = 0.999$) \shortcite{kingma_adam_2015} for 50 epochs with batch size 128. The model with the highest validation accuracy is retained for test evaluation. All models share the same initialization seed ($= 42$).

\subsection{Sign Concordance (FC Layers)}
\label{sec:supp_sign}

\begin{figure}[H]
\centering
\includegraphics[width=\columnwidth]{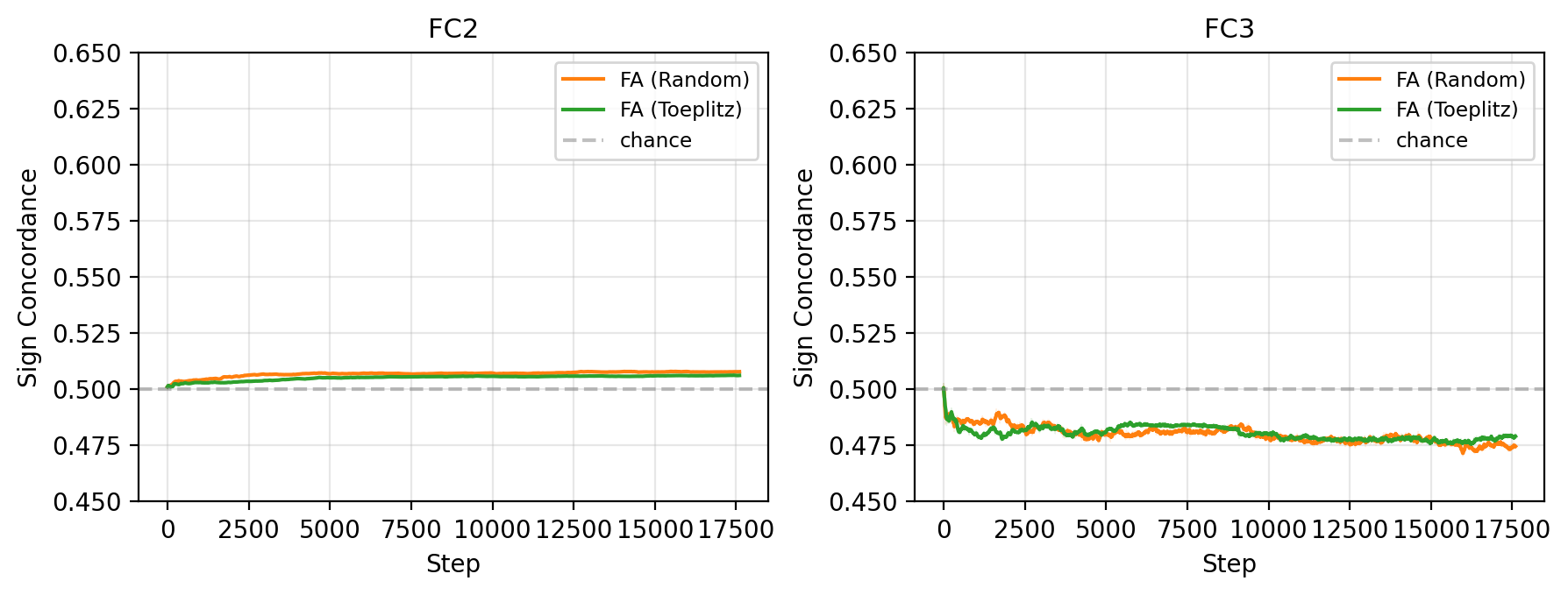}
\caption{Sign concordance for FC2 and FC3 layers during training.}
\label{fig:sign_supp}
\end{figure}

Figure~\ref{fig:sign_supp} shows sign concordance for the deeper fully-connected layers (FC2, FC3), complementing the Conv2 and FC1 results in the main text. The same pattern holds: sign concordance remains at or near the 0.5 chance level for both FA~(Random) and FA~(Toeplitz) throughout training.

\subsection{Centered Kernel Alignment: Additional Data Subsets}
\label{sec:supp_cka}

\begin{figure}[H]
\centering
\includegraphics[width=0.49\columnwidth]{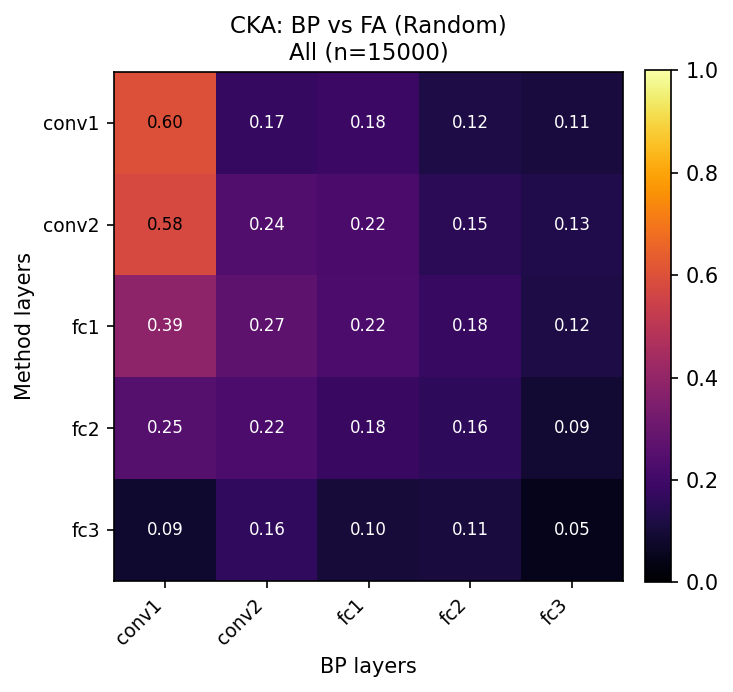}
\includegraphics[width=0.49\columnwidth]{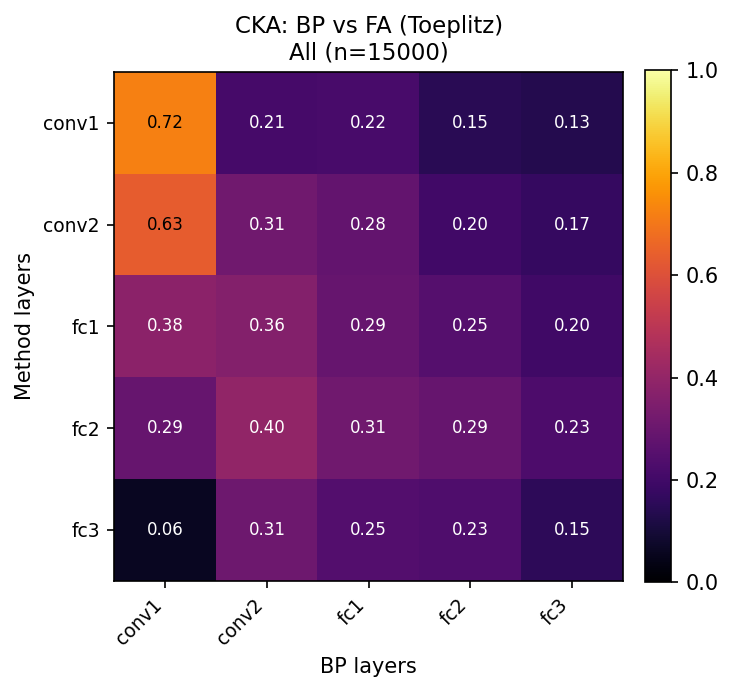}\\[2pt]
\includegraphics[width=0.49\columnwidth]{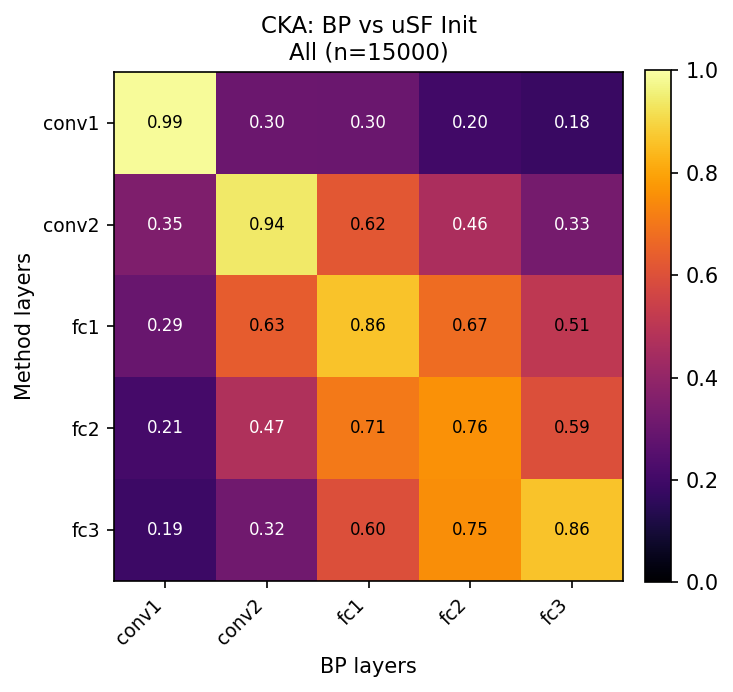}
\includegraphics[width=0.49\columnwidth]{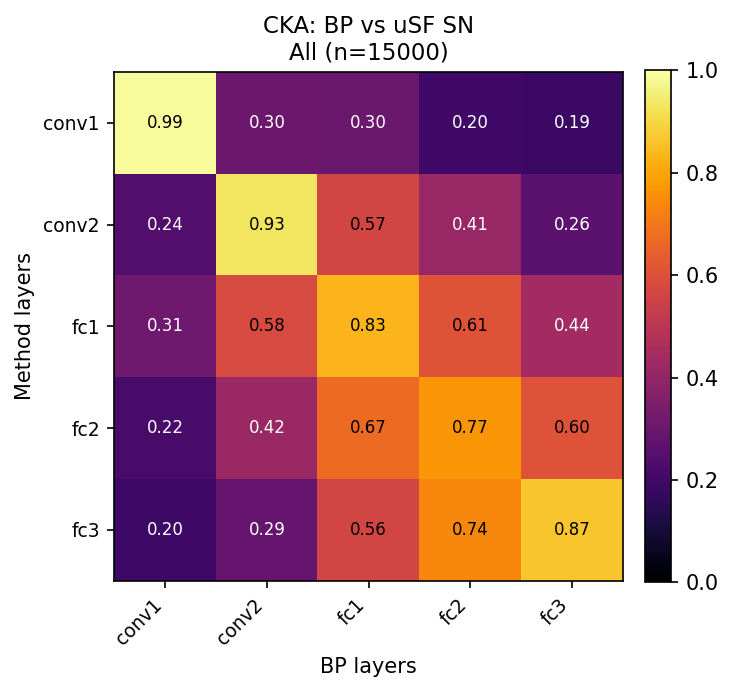}
\caption{CKA similarity (BP vs.\ each FA variant) on all validation and test samples.}
\label{fig:cka_all}
\end{figure}

\begin{figure}[H]
\centering
\includegraphics[width=0.49\columnwidth]{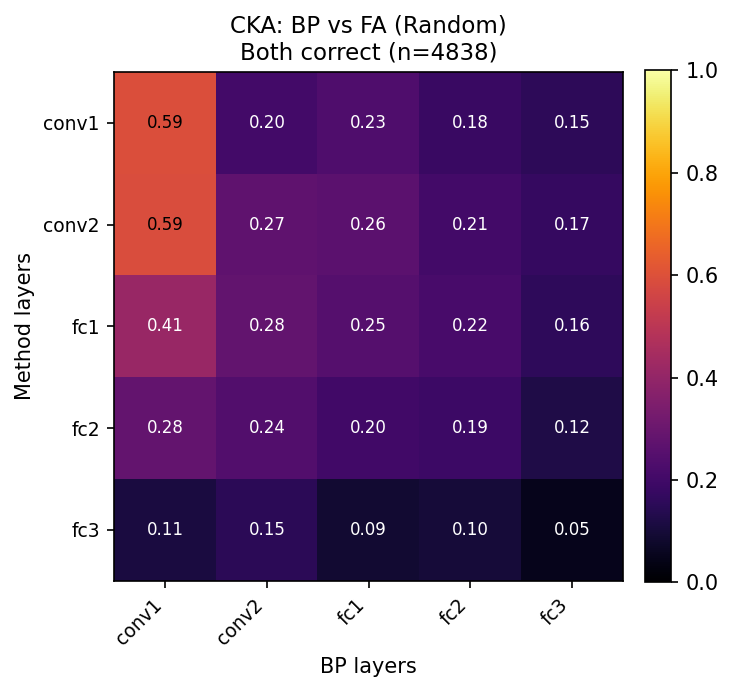}
\includegraphics[width=0.49\columnwidth]{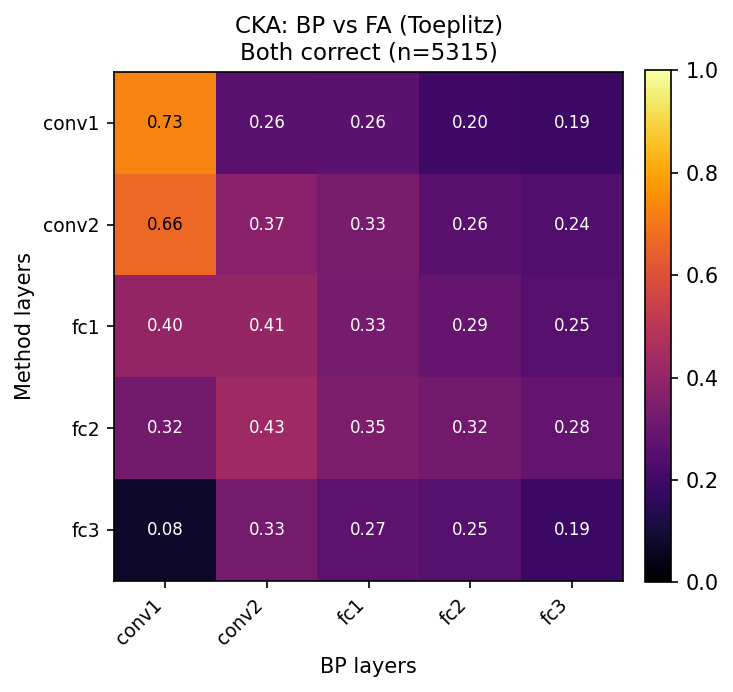}\\[2pt]
\includegraphics[width=0.49\columnwidth]{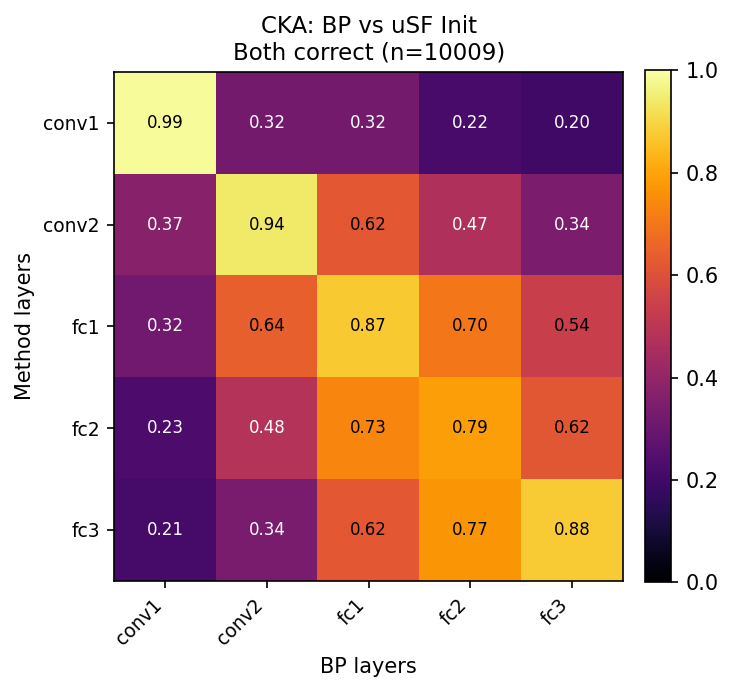}
\includegraphics[width=0.49\columnwidth]{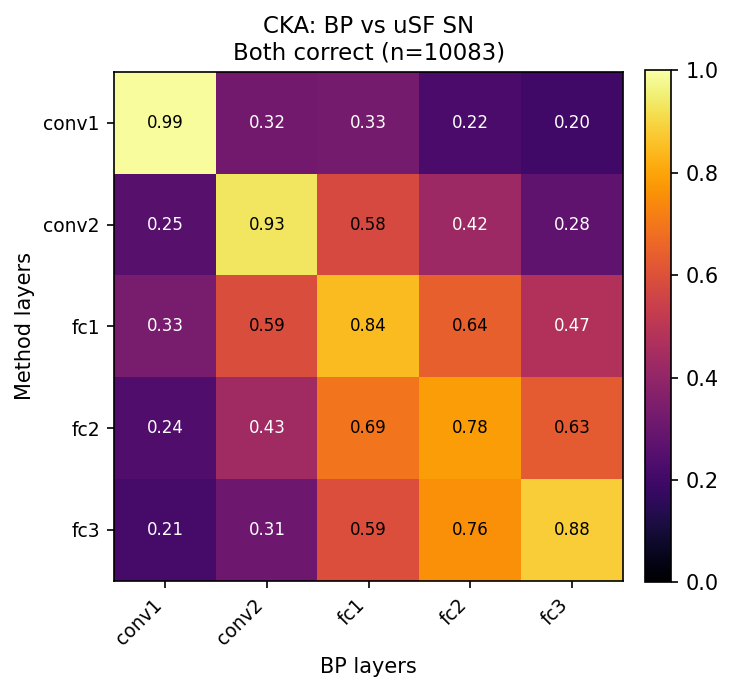}
\caption{CKA similarity (BP vs.\ each FA variant) on samples where both methods classify correctly.}
\label{fig:cka_both}
\end{figure}

Figures~\ref{fig:cka_all} and~\ref{fig:cka_both} show CKA heatmaps for the remaining data subsets described in Section~\ref{sec:cka}: all samples and the both-correct subset, respectively. Both subsets show the same qualitative pattern as the BP-correct subset in the main text: the uSF variants maintain high layer-matched similarity to BP, while the pure FA variants diverge substantially at all layers.

\end{document}